\pdfoutput=1

\documentclass[11pt]{article}

\usepackage{ACL2023}
\usepackage{subcaption}
\usepackage{graphicx}
\usepackage{times}
\usepackage{latexsym}
\usepackage{amsfonts}
\usepackage[normalem]{ulem}
\usepackage[T1]{fontenc}
\usepackage{graphicx}
\usepackage[export]{adjustbox}
\usepackage{booktabs}
\usepackage{amsmath}
\usepackage[utf8]{inputenc}
\usepackage{placeins}
 \usepackage{afterpage}
\usepackage{microtype}
\usepackage{inconsolata}

\title{Counterfactuals of Counterfactuals: a back-translation-inspired approach to analyse counterfactual editors}

\author{Giorgos Filandrianos\textsuperscript{1}
  Edmund Dervakos\textsuperscript{1}
  Orfeas Menis-Mastromichalakis\textsuperscript{1} \\
  \textbf{Chrysoula Zerva\textsuperscript{2,3}}\and
  \textbf{Giorgos Stamou\textsuperscript{1}} \\ 
  \textsuperscript{1}National Technical University of Athens \\
  \textsuperscript{2}Instituto de Telecomunicações \\
  \textsuperscript{3}Instituto Superior Técnico \& LUMLIS (Lisbon ELLIS Unit) \\
  \texttt{\{geofila, eddiedervakos\}@islab.ntua.gr, menorf@ails.ece.ntua.gr} \\
  \texttt{chrysoula.zerva@tecnico.ulisboa.pt, gstam@cs.ntua.gr}
  }

\begin{document}
\maketitle
\begin{abstract}
In the wake of responsible AI, interpretability methods, which attempt to provide an explanation for the predictions of neural models have seen rapid progress. In this work, we are concerned with explanations that are applicable to natural language processing (NLP) models and tasks, and we focus specifically on the analysis of counterfactual, contrastive explanations. We note that while there have been several explainers proposed to produce counterfactual explanations, their behaviour can vary significantly and the lack of a universal ground truth for the counterfactual edits imposes an insuperable barrier on their evaluation. We propose a new back translation-inspired evaluation methodology that utilises earlier outputs of the explainer as ground truth proxies to investigate the consistency of 
explainers. We show that by iteratively feeding the counterfactual to the explainer we can obtain valuable insights into the behaviour of both the predictor and the explainer models, and infer patterns that would be otherwise obscured. Using this methodology, we conduct a thorough analysis and propose a novel metric to evaluate the consistency of counterfactual generation approaches with different characteristics across available performance indicators.\footnote{Data and Code available at: \url{https://github.com/geofila/Counterfactuals-of-Counterfactuals}}
\end{abstract}

\section{Introduction}
\begin{figure}[t!]
    \centering
    \includegraphics[trim = 0.5cm 1.5cm 2cm 0cm , clip, width=\columnwidth]{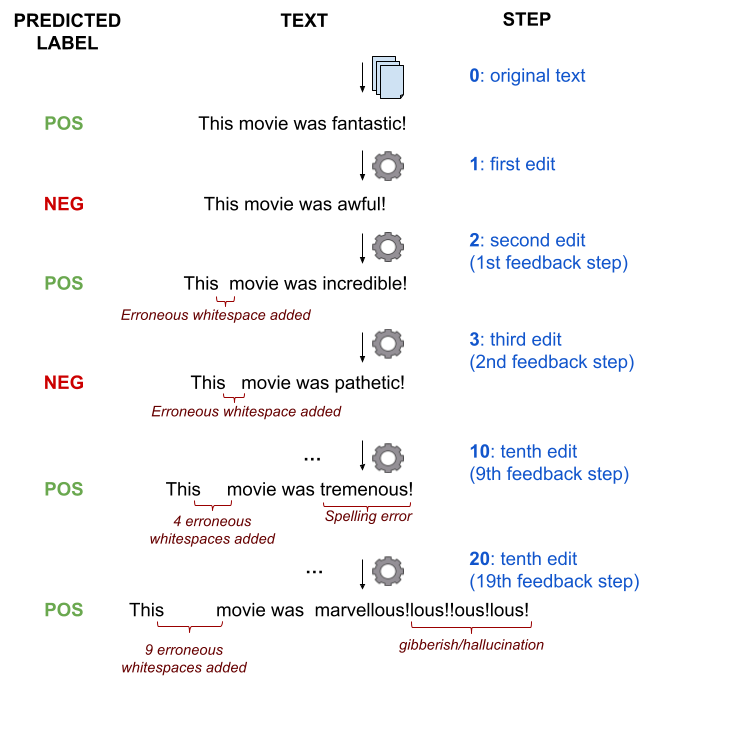}
    \caption{Using the back-translation framework to feed back the edited text to MiCE: We see the evolution of edits (centre) and predicted labels (left) through multiple feedback steps (right). As feedback steps increase, we observe an amplification of erroneous edits.}
    \label{fig:backnforth-example}
\end{figure}
The eXplainable AI (XAI) field has risen to prominence in recent years, spurred on by the success of opaque (black-box) deep learning models, which despite their impressive performance cannot be used in practice in many cases due to ethical and legal \cite{goodman2017european} concerns. To mitigate this, multiple explanation methodologies have been proposed for different tasks, data domains and use-cases \cite{bodria2021benchmarking}.  
One family of methods for explaining classifiers are \emph{counterfactual} or \emph{contrastive} explanations \cite{DBLP:journals/corr/abs-2010-10596,DBLP:journals/access/StepinACP21, balkir2022challenges}. These answer the question ``What should change in an input sample for it to be classified to class B instead of class A'' and can be especially useful in real-world applications as they offer recourse to a user. For example someone who was declined a loan by a bank's AI could ask "what is the smallest change I should do for my loan to be approved" \cite{DBLP:journals/corr/abs-1711-00399}. 

However, many XAI methods have come under scrutiny, as they can often be misleading and cause more problems than they solve \cite{rudin2019stop}.
Specifically counterfactual explainers have been shown to suffer from issues of robustness \cite{NEURIPS2021_009c434c,rawal2020can,delaney2021uncertainty}, and while there have been attempts to formalise notions of interpretability and its evaluation \cite{doshi2017towards}, there is no agreement on what constitutes a good explanation \cite{lipton2018mythos}. This ambiguity regarding the desiderata for a good explanation and the fact that they can vary according to the use-case, transcends to the chosen evaluation methods used, which vary accordingly and offer a limited perspective of the editor(s) behaviour.
Thus it is imperative that we develop further ways to evaluate and compare XAI methods in more depth.

For counterfactual explanations, existing metrics enable comparisons between different methods, however the absence of a ground truth does not allow us to assess the quality of a single explainer in a standalone mode and evaluate its output on how close it is to a (theoretically attainable) ideal explanation. Finding such an ideal explanation in practice is not easy, but we can expand on the idea of evaluation by comparison, and compare a counterfactual system's performance against itself. 
Following this rationale, we propose a methodology that 
is inspired by back-translation, which has been used for evaluating and improving machine translation systems \cite{tyupa2011theoretical, edunov2018understanding}. Specifically, by feeding the output of the system to itself (a counterfactual of a counterfactual), we would expect the result to be ``at least as good as'' the original input since we know that the original input exists, is actionable, and feasible, thus it constitutes a ``lower bound'' for the generated edit, and a proxy for ground truth. 

This methodology can be applied to obtain a lower bound on several metrics; in this work we focus on its application to \emph{minimality} since it is the metric that most of the editors attempt to minimise as a primary criterion \cite{guidotti2022counterfactual}. One of the desired characteristics of counterfactual explanations is that they constitute minimal changes to the input sample, and minimality is the metric used to measure the distance between the original and the edited samples. Due to the lack of an ideal explanation to be used as ground truth, there is no way to know if a specific value of minimality is optimal (and generally good or bad). Given a ground truth explanation, it is possible to calculate the optimal minimality to be obtained, but without it, it is only possible to compare minimality values across different edits/editors.

Using our proposed methodology we introduce a metric (which we call \textit{inconsistency}) that uses the editor's previous outputs as reference points to evaluate an editor's capability to produce minimal edits. We feed the output of the editor back to it as input to produce a new edit, and we expect the new edit to be at least as good as the edit of the previous step. For example, in Figure \ref{fig:backnforth-example} we see the outputs through different steps of our feedback loop approach, so when we feed the first edit (``This movie was awful!'') back to the counterfactual system, we expect the generated edit to be at least as good as the original text (``This movie was fantastic!''). However, we see that the editor adds an erroneous whitespace to the generated edit (see 2:second edit in Figure \ref{fig:backnforth-example}) so we know that there exists a better output that the system was not able to find, thus for sure the system did not produce the optimal output. Note that a counterfactual system with a non-zero value of inconsistency is guaranteed to be sub-optimal, however a zero value does not indicate that it is optimal. 
This approach essentially sets a lower bound for the editor, but we do not have access to the higher bound which could be impossible to define automatically. In the rest of the paper we provide a detailed description of our proposed methodology and novel metric and demonstrate its application on several frequently used editors with different characteristics.

\section{Background}
In this paper we are concerned with systems that attempt to minimally edit a given textual input so that they change the prediction of a classifier; we will henceforth refer to such systems as \emph{counterfactual editors}, or simply \emph{editors}. Below, we provide a categorisation of such systems, along with related literature, in addition to an overview of how they are being evaluated in related works.
\subsection{Counterfactual Editors}
 The intended use-case of editors varies, as do the methodologies they use for achieving their goal. For example, approaches such as MiCE \cite{ross2020explaining}, and DoCoGen \cite{calderon2022docogen} optimise their edits directly on the output of a given predictor, $g()$ by pseudo-randomly masking words in the text and optimising the proposed replacement to change the output of $g$. 
On the contrary, editors such as Polyjuice \cite{wu2021polyjuice} aim to identify generic text perturbations that can change the semantics of a sentence, without targeting a specific predictor. They frame this as \emph{general purpose counterfactuals} since they can be used for a wider range of purposes, from data-augmentation to producing counterfactual explanations or conditioned to a specific task/dataset.  
Finally, a large family of editors aim to generate \emph{adversarial examples}, whose intended use-case is to identify vulnerabilities of a classifier and expose them. Adversarial models may differ from other counterfactual editors in that they do not necessarily aim to generate a minimal or fluent edit of the original input, hence the edits might include addition of noise, etc. 
A collection of adversarial example generators for NLP including TextFooler \cite{jin2020bert} and Bert-Attack \cite{bertAttack}, is implemented in the TextAttack framework \cite{morris2020textattack}. The intuitively simpler form of such methods concerns using gradient-descent on the instance to generate examples that alter the predictor's class while simultaneously optimising the value of one or more metrics \cite{mothilal2020explaining}. Instead of attempting random permutations to generate counterexamples, other editors alter only the important features of each text. This importance is calculated in a variety of ways, including training a classifier to extract the correlation of each term with the task \cite{wang2021robustness},  measuring the effect of a feature deletion on the prediction of the classifier \cite{jin2020bert}, or using the predictor's attention \cite{ross2020explaining}. Then the important terms can be replaced with synonyms, antonyms, important terms from other tasks or using pre-trained seq2seq models
\cite{madaan2021generate, ross2021tailor,  ross2020explaining, wu2021polyjuice, fern2021text}. In our experiments, we employed editors with different intended use-cases and internal logic, namely MiCE, Polyjuice and TextFooler.

\subsection{Evaluation of Counterfactual Editors}
\label{back-metrics}

 A practical criterion for evaluating editors measures how often the output of a predictor flips to the desired class, referred to as flip-rate, validity, fidelity or attack success rate. 
Other metrics relate to the quality of generated text and include fluency, as used for example in MiCE, Polyjuice , and CAT \cite{chemmengath2021let}, grammaticality and semantics as defined in Textattack and perceptibility as defined in counterfactualGAN \cite{robeer2021generating}. These metrics rely on the use of language models, either comparing masked language model loss between original and edited text, or computing a semantic similarity between original and edited text. 
Proximity as described in \cite{DBLP:journals/corr/abs-2010-10596,keane2021if} refers to generic distance measures between edited and original data. For natural language processing, the distance metric used is typically the word level Levenshtein edit distance \cite{levenshtein1966binary}, referred to also as minimality (MiCE), closeness (Polyjuice) and edit distance (CAT). 
There are more criteria that have been used for the evaluation of counterfactuals, such as sparcity \cite{keane2021if} referring to the number of features being changed, and closeness to the training data \cite{DBLP:journals/corr/abs-2010-10596} or relative distance \cite{keane2021if} involve comparing the explanations with instances from the training data. 
Finally, a more recent approach for evaluating counterfactual explanations involves measuring the degree to which the explanations help a student model \cite{pruthi2022evaluating} or a human \cite{doshi2017towards,treviso2020explanation} to learn to simulate a black-box teacher model. This provides a measure of informativeness of the explanations. 

In this work we focus on evaluation with automated metrics that do not require human input or external data, and are most frequently used in editor evaluation, namely, minimality, flip-rate and fluency. These metrics aim to quantify different aspects of editors' behaviour in a comparative fashion. Instead, we propose an inconsistency metric which allows to set a lower bound on the editor's ability to reach an optimal counterfactual and as such can be useful without the need to compare to other editors. For instance, a minimality value for a given editor carries no information on its own regarding optimality. However, if an editor has a value $inc@1 = x$ for the inconsistency of minimality, it means that on a given test-set it missed the optimal counterfactual solution at least by an average of $x$ tokens.

\section{Back-translation for analyzing editors}
\label{sec:backtranslation}

We formalise our problem as follows. We assume access to a  classifier $g$ such that  $g:\mathcal{L}\rightarrow{[0,1]}^C$, where $\mathcal{L}$ the set of text for a specific language and $C$ is the number of different classes. We then consider the  counterfactual editors for $g$ as functions $f:\mathcal{L}\rightarrow{\mathcal{L}}$, and we assume that the goal of the editor $f$ is threefold:

\begin{enumerate}
    \item The edited text is classified to a different class  $\arg\max{g(f(x))}\neq{\arg\max{g(x)}}$.
    \item The edits are minimal with respect to some distance metric $d$: $f=\arg\min_{h\in{\mathcal{F}}}{d(x,h(x))}$, where $\mathcal{F}$ is the set of functions for which $\arg\max{g(f(x))}\neq{\arg\max{g(x)}}$.
    \item The edited text $f(x)$ is fluent and within the distribution of $\mathcal{L}$. 
\end{enumerate}

To examine the degree to which these criteria hold, we analyse the behaviour of editors when they are iteratively fed back with their output, i.e., we are studying the function $f(f(f(...f(x))))$, and evaluating the three criteria described above after $n$ applications of the editor. Specifically, we first define a novel evaluation metric to quantify the second criterion based on the iterative feedback approach, and then we discuss how the first and third criteria can be more thoroughly checked by measuring performance metrics after $n$ steps of feedback (notated as $\mathsf{metric}@n$).
\subsection{(In)consistency of minimality}
Intuitively, since the edits are ideally minimal, if a sentence $A$ is edited into sentence $B$ and their distance is $d(A,B)$, then feeding back sentence $B$ to the editor should yield a sentence $C$ for which $d(B,C)\leq{d(A,B)}$, otherwise $C$ is not the result of a minimal edit. This inequality holds based on that (a) we know that $A$ exists, (b) we assume all textual edits to be reversible, hense $A$ is reachable from $B$ and (c) $d$ is symmetric, meaning $d(A,B)=d(B,A)$. Thus, in this case, $A$ can be used as a proxy to a ground truth, to be compared with $C$. 
Given a distance metric $d$ (such as Levenshtein distance, embedding cosine similarity, etc.), we can measure how consistent the counterfactual editor is w.r.t $d$ by iteratively feeding back the edited text to the editor and measuring the change in the value of $d$. 
Specifically, given an editor
$f:\mathcal{L}\rightarrow{\mathcal{L}}$, a text
$x\in{\mathcal{L}}$ and a distance 
$d:\mathcal{L}\times{\mathcal{L}}\rightarrow{\mathbb{R}^+}$ we define the \emph{inconsistency of $f$ with respect to $d$, for $x$ as:}.
\begin{equation}
    \mathsf{inc}(f,x)=\mathsf{relu}[d(f(f(x)),f(x))-d(f(x),x)] .
    \label{eq:inc}
\end{equation}

The difference $d(f(f(x)),f(x))-d(f(x),x)$ shows how much the distance $d$ changes between consecutive applications of the editor $f$ and the $\mathsf{relu}$ function allows to take into account only the increase of the distance. This is important, because a decrease in the distance, which would correspond to a negative difference, is not necessarily an indicator of a good set of edits. It could, for example, indicate that not enough changes were made, and there is no way to know if that is the case, or if a better, more minimal set of edits was found. Contrarily, when the value is positive, we have a \emph{guarantee} that a better set of edits exists, namely, the one of the previous feedback step. Equation \ref{eq:inc} counts the difference in $d$ after a single feedback iteration though the editor, but as with other metrics in this work, we can keep feeding back the output of the editor to itself, and compute $\mathsf{inc}(f,f(x))$ to get more information about the editor's inconsistency. When we do this, we measure the average inconsistency after $n$ steps of feedback as:
\begin{equation}
    \mathsf{inc}@n(f,x)=\frac{1}{n}\sum_{i=0}^{n-1}{\mathsf{inc}(f_{i+1}(x),f_i(x))},
    \label{eq:incn}
\end{equation}
where $f_0(x)=x$ and $f_i(x)=f(f_{i-1}(x))$.

\subsection{Diverging from the distribution}
Of the desiderata for counterfactual editors, the constraint that $f(x)$ is fluent and within distribution can be hard to verify, as the true distribution of texts $\mathcal{L}$ may be inaccessible, 
and fluency is hard to evaluate in a systematic and automated way. 
The token-level perplexity of a large language model is a frequently used proxy to the fluency estimation, employed in multiple NLP tasks \cite{john2018disentangled,wang2019bert,vasilakes2022learning}. It involves using a language model $\mathcal{M}_\mathcal{D}$ trained on a large dataset $\mathcal{D}$ and computing the averaged perplexity over a given sequence of text $x = {x_1, x_2, ..., x_T}$ as follows: 
\begin{equation}
    \mathrm{PPL}(x) = \exp{\biggl\{ \frac{1}{T} \sum_{t=1}^T \log{p_{\mathcal{M}_\mathcal{D}}(x_t | x_{1:t-1})} \biggl\} }.
    \label{eq:ppl}
\end{equation}

Note that for the fluency estimation $\mathcal{M}_\mathcal{D}$ in equation \ref{eq:ppl} is not finetuned on $\mathcal{L}$. Assuming $\mathcal{L}$ is accessible, we can also fine-tune $\mathcal{M}_\mathcal{D}$ on it, obtaining a model $\mathcal{M}_\mathcal{L}$ that we can use to detect out-of-distribution (OOD) cases \cite{arora2021types} using the same $\mathrm{PPL}$ formula described in equation \ref{eq:ppl}, since in this case we can assume that the probability over each token predicted by $\mathcal{M}_\mathcal{L}$ reflects the probability under the distribution of $\mathcal{L}$. We employ and compare both $\mathrm{PPL}$ over  $\mathcal{M}_\mathcal{D}$ ($\mathrm{PPL}_{\mathcal{D}}$) and $\mathrm{PPL}$ over  $\mathcal{M}_\mathcal{L}$ ($\mathrm{PPL}_{\mathcal{L}}$) in our experiments. Furthermore,  $\mathrm{PPL}_{\mathcal{L}}@n$ and $\mathrm{PPL}_{\mathcal{D}}@n$ could more clearly show the divergence from the distribution of generated samples, and the deterioration of fluency. 



\subsection{Flip rate}
So far we have not taken under consideration the ability of the editor to change the predicted class, which is typically measured as flip rate. Many recently proposed counterfactual editors achieve flip rates above $95\%$, leaving a small margin to confidently compare editors with respect to this metric. Using the proposed feedback methodology, we can measure the flip rate after $n$ loops of feedback to get more detailed information about the ability of the editor to consistently change the predicted class after several perturbations of the sentence. 

\section{Experimental Setup}
\label{sec:setup}
We evaluate our approach on two different datasets and classifiers trained on them.
Specifically, we use a binary classifier trained for sentiment classification on the IMDb dataset \cite{maas-EtAl:2011:ACL-HLT2011} and a multi-class short-document classifier trained on Newsgroups \cite{lang1995newsweeder}.  We provide details and statistics for each dataset in Appendix \ref{appendix:dataset}.

Using these classifiers, we 
apply our methodology on the three counterfactual editors described in \S\ref{sec:editors} and examine the metrics described in \S\ref{sec:backtranslation} by generating edits, testing the classifiers on them and feeding back the edited text to the editor for $n=10$ steps. For each editor we apply our methodology as follows: for a given input text $x$, at step $n$ we select from the pool of generated edited texts the one with the minimum minimality that alters the prediction, if such an output exists, otherwise we select the minimum minimality output. We repeat this process until $n=10$ and discuss the behaviour of each editor across metrics in \S\ref{sec:inc-results} and \S\ref{sec:otherMetrics}.



We also study the impact of test-set size on the observed differences between results and the statistical significance of findings. We present detailed results in Appendix \ref{appendix:t_test}. We found that for a test set size greater than 200 texts results converge on both datasets and we obtain statistically significant differences. Based on these findings we randomly sample 500 texts from the IMDb dataset for our experiments, to reduce the computational load. Since NewsGroups is smaller, we use the full dataset.

\subsection{Evaluation}
The metrics that we use with our methodology are:

\noindent\textbf{Minimality:} The word-level Levenshtein distance between original and edited text.

\noindent\textbf{inc@n:} Inconsistency of word-level Levenshtein distance as per equation \ref{eq:incn}.

\noindent\textbf{Flip rate:} The ratio $\frac{\mathsf{n_{flipped}}}{\mathsf{n_{all}}}$, where $\mathsf{n_{all}}$ is the size of the dataset, and $\mathsf{n_{flipped}}$ 
are the samples for which the prediction changes after applying the editor.



\noindent\textbf{ppl-base:} Language model perplexity  of GPT-2, a large, general-domain language model \cite{radford2019language}, as per equation \ref{eq:ppl}.

\noindent\textbf{ppl- fine:} Language model perplexity  of GPT-2, fine-tuned on IMDB \footnote{https://huggingface.co/lvwerra/gpt2-imdb} and on Newsgroups \footnote{https://huggingface.co/QianWeiTech/GPT2-News}. Used to examine how ``unexpected'' the edited text is with respect to each dataset.

We also compute the above metrics after $n$ steps of feedback, with the exception of inc@n, which uses the feedback steps by definition. 



\subsection{Editors}
\label{sec:editors}
We experimented with three editors with different characteristics. Brief descriptions of each editor and the main differences between them are presented below, and more  details in Appendix \ref{appendix:exp-set}.  

\paragraph{Polyjuice} 
Polyjuice is a general-purpose counterfactual generator that produces perturbations based on predefined control types. This editor employs a GPT-2 model that was fine-tuned on various datasets of paired sentences (input-counterfactual), including the IMDb dataset. Polyjuice does not use the classifier predictions during the generation of the counterfactual texts but rather focuses on the variety of the edits based on a learned set of control codes such as "negation" or "delete". 

\paragraph{MiCE}
MiCE is a two-step approach to generating counterfactual edits. It uses a T5 deep neural network to fill the blanks on a text, which is fine-tuned in the first step to better fit the dataset distribution. In the second step, the input text is masked either randomly or by using predictors attention in a white box manner, and the fine-tuned model fills these blanks. This step aims to learn the minimum edits that will alter the classifier's prediction. In our experiments, we employed MiCE in a white box manner, meaning that the fine-tuning is done by using the predictor's outputs (and not the ground-truth label), and for selecting the masks' locations, the classifier's attention is used in order to compare its result with Polyjuice which also utilised a deep neural network for the generation, but uses the predictor in a black box manner.

\paragraph{TextFooler}
TextFooler is a method for generating adversarial examples for a black-box classifier. The generation process differs from other editors since it does not employ a deep neural network, such as GPT2 or T5, to construct the produced counterfactual. Instead consists of finding the words that are influential to the predicted class and replacing them in a more deterministic way. The influence of each word is calculated by measuring how its removal alters the predictor's output \cite{liang2017deep}. The alternatives that can replace a word are predefined to be the closest match in the embedding space and are thus independent of the rest of the sentence and the classifier. Hence, TextFooler chooses single word replacements that are synonyms with the removed word, with the added constraint of having the same part-of-speech.

\section{Interpreting the inc@n metric}
\label{sec:inc-results}

In Table \ref{tab:inc} we show the results of the proposed inc@n metric. 
It is important to mention that there is an intuitive interpretation of the inc@n metric. 
Since we use Levenstein distance in our experiments, inconsistency corresponds to the mean number of tokens that the editor is altering on top of those that were needed to produce a valid counterfactual. The reasons behind inconsistency could vary depending on the mechanism of selecting important parts of the source text, the generation procedure, or the algorithm for locating the best edits.

We can observe significant differences for inc@n 
between the editors, reflecting the differences in the underlying approach. TextFooler is the most consistent editor, with low inc@n values that imply very rare increases in minimality between steps. This shows that the 
controlled nature of TextFooler's approach for selecting 
replacements is beneficial to the generation of 
consistent explanations. MiCE and Polyjuice on the other hand are less consistent, which could be attributed to the use of large language models in the generation process, which 
are more sensitive to small perturbations that can 
alter their output \cite{jin2020bert}. 

The comparatively high value of inconsistency for Polyjuice at the first steps for IMDb, can be explained by the fact that it has to ``guess'' the locations it should change in the text without access to the predictor. Especially for longer inputs,  the search space of Polyjuice is exponentially larger. This forces it to make more aggressive edits to achieve the same result, often deleting a large portion of the source text and seems to contribute to Polyjuice's reported robustness issue \cite{madsen2022post}. 
For example, Polyjuice erased over 70\% of the original text for 83\% of the first two steps of edits on the IMDb dataset  (candidates with lower minimality may be produced but failed to change the class and were rejected). An extensive analysis of this tendency across all editors and datasets is presented in the Appendix \ref{sec:prunning}. 
This ``extreme erasure'' pattern disappeared in the next steps, where the input length was significantly smaller (for instance original texts have on average 204 tokens, while the edited ones produced by Polyjuice have 29).
On the other hand, for shorter texts, for which the search space is smaller, Polyjuice is more consistent, without radical changes on the original input. We hence attribute the differences between the first and later steps of the $inc@n$ metric to this tendency of Polyjuice to reduce the length of long texts. After the first step since the length of the input texts have already been significantly reduced, its behaviour is more consistent. This pattern is invariant between the two datasets but not visible on the first steps of Newsgroups results since it contains texts with 43\% fewer tokens than IMDb on average.



\begin{table}[t]
\centering
\caption{Inconsistency (inc@n) computed on the IMDb and Newsgroups datasets.}
\scalebox{0.8}{
\begin{tabular}{cccc}
\toprule
                           & MiCE  & Polyjuice & TextFooler \\ \cline{2-4} 
                           & \multicolumn{3}{c}{IMDb}       \\ \hline
\multicolumn{1}{c|}{inc@1  $\downarrow$} & 0.86  & 6.21      & 0.01    \\ 
\multicolumn{1}{c|}{inc@2 $\downarrow$} & 5.95 &	4.65 &	0.33      \\
\multicolumn{1}{c|}{inc@3  $\downarrow$} & 4.65  & 3.98      & 0.36       \\
\multicolumn{1}{c|}{inc@5  $\downarrow$} & 4.87  & 2.9       & 0.47       \\
\multicolumn{1}{c|}{inc@9  $\downarrow$} & 4.73  & 2.22      & 0.49       \\  \hline
\multicolumn{1}{c|}{}      & \multicolumn{3}{c}{Newsgroups} \\ \hline
\multicolumn{1}{c|}{inc@1  $\downarrow$} & 11.11 & 0.99      & 0.04  \\
\multicolumn{1}{c|}{inc@2  $\downarrow$} & 7.97 &	1.29 &	0.55  \\
\multicolumn{1}{c|}{inc@3  $\downarrow$} & 7.89  & 1.35      & 0.46       \\
\multicolumn{1}{c|}{inc@5  $\downarrow$} & 6.92  & 1.3       & 0.49       \\
\multicolumn{1}{c|}{inc@9  $\downarrow$} & 6.11  & 1.21      & 0.46     \\
\bottomrule
\end{tabular}
}
\label{tab:inc}
\end{table}

To better understand what these values of inc@n represent, in Figures~\ref{fig:all-mini}, \ref{fig:all-inco} we show box-plots of minimality and inc@n after each step of feedback for the IMDb dataset (box-plots for Newsgroups appear in Appendix \ref{ap:news}). While the tendency is for minimality to decrease, implying that the editors tend to perform fewer edits after each feedback step, there are cases where it increases, implying inconsistency. 

\begin{figure}[b]
    \centering
    \includegraphics[scale=0.20, frame]{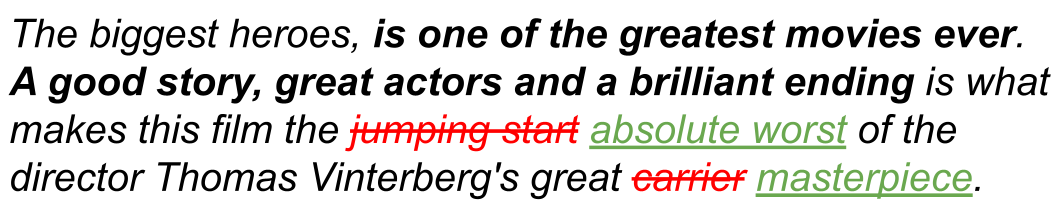}
    \caption{MiCE example of an IMDb dataset sample. }
    \label{fig:ex-remenants}
\end{figure}

In Figure~\ref{fig:all-inco}, what stands out is the higher value of inc@n for MiCE when $n$ is even. Since even steps correspond to the attempt to move from the original class to a different one, higher inc@n indicates that it is easier to transition to the original class than from it. This could be attributed to remnants of the original input text pushing the classifier towards the original prediction, and thus requiring fewer edits. 
For instance in Figure~\ref{fig:ex-remenants} we can see an example of an edit produced by MiCE, where clearly parts of the original text indicating positive sentiment (marked in bold) were left unchanged. 
On Newsgroups, MiCE is significantly more inconsistent especially in the first steps, probably due to its requirement to select a specific target class in contrast to the other editors (see also Appendix \ref{ap:news}).  

These observations highlight the need for feedforward evaluations of such systems since the minimality@1 reveals only a limited, dataset-dependent aspect of editors' capabilities and performance. Furthermore, they show the effectiveness of additional feedback steps to more accurately quantify the difference between the samples produced by an editor, and obtain a proxy for a global minimum. 
In Table \ref{tab:inc} we see that from $inc@3$ on the obtained inc@n values start to converge for both datasets.

It is important to mention that the inconsistency of minimality captures different attributes of the editor than the minimality itself. High minimality means that the editor made more edits in order to alter the label of the input text. This may be due to either a weakness of the editor or the input itself requiring more edits to change class. To exclude the latter hypothesis, it is necessary to find counterfactual examples with lower minimality than the one produced, to confirm that there are better states that the editor could not explore. However, these states must meet the exact same conditions that the editor takes into account. The three editors analysed in this study have a secondary objective of producing realistic counterexamples; hence, for instance, the addition of random characters in the middle of the text is not a desirable goal state, despite the fact that it may result to a label flip with lower minimality. Along the same lines, TextFooler aims to replace every word with a synonym; therefore, a counterexample that replaces a word with an antonym (e.g., 'love' with 'hate') is not an acceptable goal state for it. To our knowledge, there have been no efficient or impartial methods for finding counterexamples with a lower value of a specified metric (such as minimality) that also meet exactly the same requirements as the editor. The proposed methodology comes to fill this gap, and the inconsistency metric can quantify the weaknesses of the editor in terms of the studied metric, in this case, minimality. In short, a positive inconsistency proves that there are goal states with a lower value of the corresponding metric that the editor should, but did not explore.

\section{Additional insights from counterfactuals of counterfactuals}
\label{sec:otherMetrics}
Besides measuring minimality and inc@n, we also investigated how the feedback approach can give us additional insights for the other two desiderata for editors, flip-rate and fluency.

\subsection{Flip Rate}
In Table~\ref{tab:flip} we show flip-rate measured after applying the feedback methodology. 
At the first step MiCE has a perfect flip rate; if analysed in solitude this observation might lead to the erroneous conclusion that the model can always alter the class of any text. However, this is a test-set-dependent result and does not apply in general since the flip rate reduces significantly in the following steps. Hence, there are instances closer to its distribution (Section \ref{sec:fluency}) in which MiCE could not alter the predicted class. Conversely, the flip-rate of Polyjuice and TextFooler increases for later feedback steps.


\begin{table}[t]
\centering
\caption{Flip-rate after feeding the original text to the editor once (@1), and after 9  steps of feedback (@9) for the IMDb and Newsgroups dataset.}
\scalebox{0.8}{
\begin{tabular}{cccc}
\toprule
                           & MiCE  & Polyjuice & TextFooler \\ \cline{2-4} 
                           & \multicolumn{3}{c}{IMDb}       \\ \hline
\multicolumn{1}{c|}{Flip Rate@1 $\uparrow$}       & \textbf{1.000}   &  0.8747   &   0.6195   \\
\multicolumn{1}{c|}{Flip Rate @9 $\uparrow$}      & 0.8561  &  \textbf{0.9675}   &   0.7865   \\  \hline
\multicolumn{1}{c|}{}      & \multicolumn{3}{c}{Newsgroups} \\ \hline
\multicolumn{1}{c|}{Flip Rate@1 $\uparrow$}       & \textbf{0.87}   &  0.77   &  0.79    \\
\multicolumn{1}{c|}{Flip Rate @9 $\uparrow$}      & 0.836  &  \textbf{0.968}   &   0.89   \\
\bottomrule
\end{tabular}
}
\label{tab:flip}
\end{table}



\label{sec:fluency}

\begin{figure*}
\label{fig:fluency}
\centering
%
\begin{subfigure}[b]{0.33\textwidth}
\includegraphics[width=\textwidth]{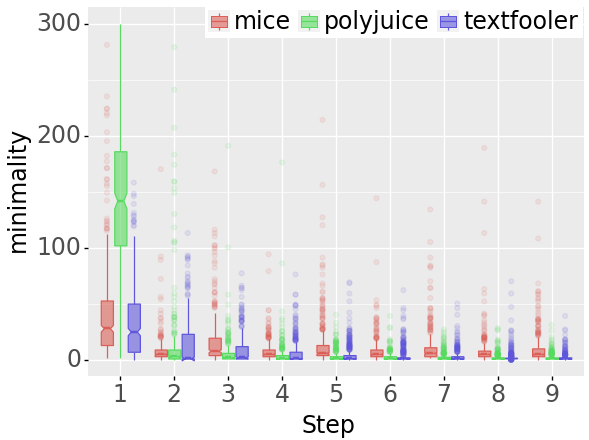}
    \caption{Minimality.}
    \label{fig:all-mini}
\end{subfigure}
\begin{subfigure}[b]{0.32\textwidth}
\includegraphics[width=\textwidth]{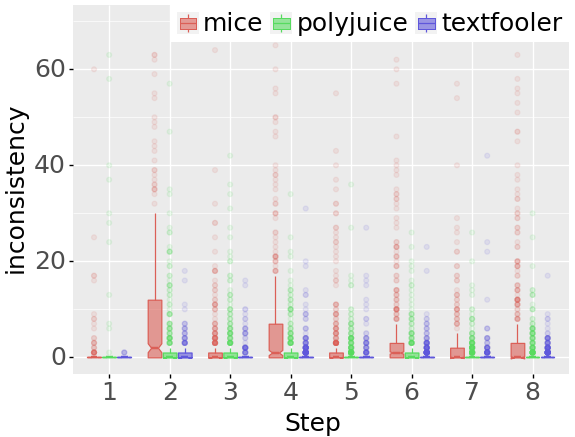}
\caption{Inconsistency of minimality.}
\label{fig:all-inco}
\end{subfigure}\\
\begin{subfigure}[b]{0.32\textwidth}
\centering
\includegraphics[width=\textwidth]{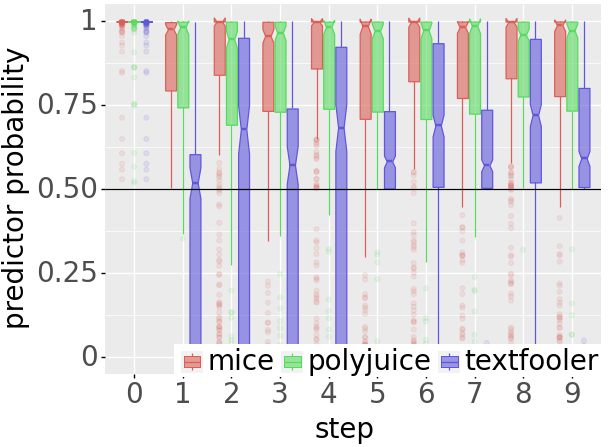}
\caption{Probability of the target class.}
\label{fig:all-probs}
\end{subfigure}
\begin{subfigure}[b]{0.33\textwidth}
\includegraphics[width=\textwidth]{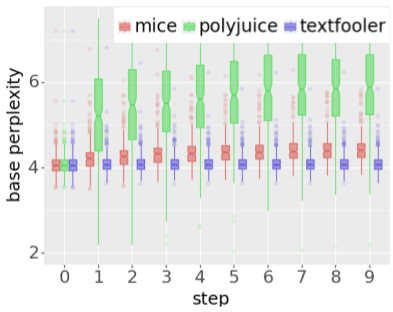}
\caption{PPL of base GPT-2.}
\label{fig:base-ppl}
\end{subfigure}
\begin{subfigure}[b]{0.33\textwidth}
\centering
\includegraphics[width=\textwidth]{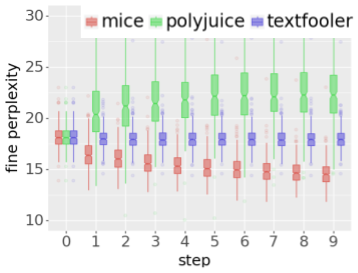}
\caption{PPL of fine  GPT-2.}
\label{fig:fine-ppl}

\end{subfigure}
\caption{Minimality, inc@n, and predictor probability, base-ppl and fine-ppl, after each step of feedback and for each editor on the IMDb dataset.}
\end{figure*}

To investigate this, in Figure~\ref{fig:all-probs} we show the prediction probabilities for the target class after applying each editor and after each step of feedback, where a sample is flipped if it has a target prediction probability greater than 0.5. 
Comparing this figure with the corresponding figure of the inconsistency of minimality (Figure~\ref{fig:all-inco}), we can observe the same patterns, such as differences in even and odd steps for MiCE. Thus this figure seems to corroborate the difficulty of the editor to return to the original class from the counterfactual.


\subsection{Fluency}

We use two metrics to approximate the fluency of the generated texts, shown in Table~\ref{tab:fluency}. 
For the ppl-base indicator, TextFooler has the lowest value, indicating fluent text, while the value does not change after several feedback steps, which further supports the editor's consistency. In contrast, MiCE's fluency slightly deteriorates after feedback, which coincides with 
the editor's inconsistency (compared to TextFooler). 
Furthermore, in Figures~
\ref{fig:base-ppl}, \ref{fig:fine-ppl} we show the evolution of the two fluency related metrics for each feedback step. TextFooler's fluency is relatively stable across both metrics, on par with its low inconsistency. On the other hand, 
Polyjuice 
appears to deteriorate in fluency, as both the base-PPL and fine-PPL indicators have an increasing trend. The most striking difference in the perplexity patterns lies between MiCE and Polyjuice, where for the base model perplexity is consistently increasing for both editors, while for the fine-tuned model, MiCE's perplexity decreases and Polyjuice's continues to increase. We can thus deduce that while both editors produce changes that have a negative impact in the overall fluency, in the case of MiCE which is trained exclusively on IMDb data, the edited texts are closer to the IMDb distribution, hinting at an ``overfitting'' behaviour of sorts. Instead, Polyjuice takes advantage of different datasets during training and produces more diverse edits.

\begin{table}[t]
\centering
\caption{Metrics for measuring fluency computed for three counterfactual editors, of the IMDb and Newsgroups datasets, after feeding the original text to the editor once (@1), and after 8 additional steps of feedback (@9)}
\label{tab:fluency}
\scalebox{0.8}{
\begin{tabular}{cccc}
\toprule
                           & MiCE  & Polyjuice & TextFooler \\ \cline{2-4} 
                           & \multicolumn{3}{c}{IMDb}       \\ \hline
\multicolumn{1}{l|}{ppl-base@1 $\downarrow$}   & 4.2546  &  7.4525   &  \textbf{4.1178}    \\
\multicolumn{1}{l|}{ppl-base@9 $\downarrow$}   & 4.4512  &  7.3825   &  \textbf{4.1161}   \\
\multicolumn{1}{l|}{ppl-imdb@1 $\downarrow$}   & \textbf{16.5315} & 33.4798   &  18.0662   \\
\multicolumn{1}{l|}{ppl-imdb@9 $\downarrow$}   & \textbf{14.6069} & 27.8074   &  17.9917   \\ \hline
\multicolumn{1}{c|}{}      & \multicolumn{3}{c}{Newsgroups} \\ \hline
\multicolumn{1}{l|}{ppl-base@1 $\downarrow$}   &  
 5.164 & 8.926  &   \textbf{4.801}   \\
\multicolumn{1}{l|}{ppl-base@9 $\downarrow$}   & 5.36  & 7.878  &   \textbf{4.776}  \\

\multicolumn{1}{l|}{ppl-newsgroup@1 $\downarrow$} & 4.27  & 6.67 &  \textbf{3.99} \\
\multicolumn{1}{l|}{ppl-newsgroup@9 $\downarrow$} & 4.4  &  5.90  & \textbf{3.98}   \\ \bottomrule
\end{tabular}
}
\end{table}

\section{Conclusion}
In this work we introduced a methodology for analysing different aspects of counterfactual editors and obtaining an approximate ground truth by iteratively feeding back their output. Combined with evaluation metrics from related literature, our proposed approach provides new ways to understand the counterfactual editors' behaviour and performance based on their intended use case, and thus help develop better editors. We proposed inc@n, a metric to measure the consistency of editors, and we showed how the proposed approach can help diagnose and analyse a diverse set of existing editors and gain new insights on their behaviour, such as those made apparent by observing the discrepancies of odd and even steps of feedback. 

Our findings allow for a more interpretable evaluation of editors that goes beyond mere comparisons between them. The results motivate further research in this direction including experiments on additional evaluation metrics, editors and tasks. 
Apart from expanding the scope of our experiments, we intend to look into using the feedback information to automatically address the weaknesses and inconsistencies of editors during fine-tuning, to obtain more robust and interpretable counterfactual edits.
Along these lines, we also plan to investigate the benefit of integrating the feedback rationale into the training of counterfactual generation algorithms; for example a back-translation inspired objective could help alleviate the problematic behaviour and boost performance.

\section*{Limitations}
This work focused on experiments on English datasets and did not explore other languages. While we expect that our assumptions hold across languages and the proposed methods and metrics can be applied without any further modifications to other languages this has not been explicitly verified. Additionally, we ensured to experiment with counterfactual editors that are representative of the main counterfactual editing methodologies, however we did not exhaustively cover all publicly available editors as our main goal was to demonstrate that our proposed method is widely applicable rather than to exhaustively compare editors.

\section*{Acknowledgements}
This work was supported by the European Research Council (ERC StG DeepSPIN 758969)  and  by the Funda\c{c}\~ao para a Ci\^encia e Tecnologia through contract  UIDB/50008/2020.

\bibliography{anthology,custom}
\bibliographystyle{acl_natbib}

\clearpage
\appendix

\section{Dataset Analytics}
\label{appendix:dataset}
\subsection{IMDb}
The original IMDb dataset consists of 50K movie reviews split evenly between positive and negative ones (binary classification). We randomly sample 500 documents from the dataset to generate a test-set for our experiments. 

In the sampled test-set, the mean number of tokens and characters of the selected comments are 204 and 1000, with a standard deviation of 112 and 562, respectively. In addition, 52\% of these comments are classified as "positive" while 48\% as "negative". 
The mean number of characters and tokens for inputs that are classified with "positive" sentiment is $990$ and $530$, with a standard deviation of $204$, and $108$, respectively. The distribution for texts that are classified with "negative" sentiment is similar, where the mean number of characters and tokens is $1006 \pm 589$ and $204 \pm 115$, respectively.

\subsection{Newsgroups}
The original Newsgroups dataset consists of 20K short documents split evenly between 20 newsgroup classes, representing the document topic. We use the test-set partition which consists of 7K documents for our experiments, as it is provided from scikit-learn library \footnote{https://scikit-learn.org/0.19/datasets/twenty\textunderscore newsgroups.html}, since the train set has already used for fine-tuning some of the editors. The mean number of characters and tokens for this dataset is 603 and 207, with a standard deviation of 495 and 103, respectively. 

The list of the 20 classes present in the dataset is: [comp.graphics, 
comp.os.ms-windows.misc, 
comp.sys.ibm.pc.hardware, 
comp.sys.mac.hardware, 
comp.windows.x, rec.autos, 
rec.motorcycles, 
rec.sport.baseball, 
rec.sport.hockey sci.crypt, 
sci.electronics, 
sci.med, 
sci.space, 
misc.forsale, talk.politics.misc, 
talk.politics.guns, 
talk.politics.mideast, talk.religion.misc, 
alt.atheism, 
soc.religion.christian]



\section{Experimental Setup}
\label{appendix:exp-set}
For both of our experiments,  we used the predictors that are used in MiCE (since MiCE requires white box access to the predictor, and we wanted to intervene as little as possible in editors' code). These predictors were built based on ROBERTA-LARGE and were fixed during the evaluation. The predictors' accuracy is the same as stated in the proposed paper, 95.9\% for IMDb and 85.3\% for the Newsgroups.  

We compare three counterfactual editors (MiCE, Polyjuice, and TextFooler) using the same classifier by making changes, testing the classifier on them, and feeding back the modified text to the editor ten times. Editors create numerous altered versions at each feedback stage and for each input text. We choose the output with the lowest minimality that changes the prediction (counterfactual goal), if such an output exists; otherwise, we choose the output with the lowest minimality. This design choice was made because there were cases in which an editor may not alter the prediction of the produced text by it made this transformation in the following steps. 

\paragraph{MiCE} For the editor of mice, we used the pre-trained T5 model that the authors provided\footnote{https://storage.googleapis.com/allennlp-public-models/mice-imdb-predictor.tar.gz} \footnote{https://storage.googleapis.com/allennlp-public-models/mice-newsgroups-editor.pth}. This model was fine-tuned on the same data as the predictor. For the generation procedure, we left the default arguments for each one of the datasets as the authors supplied on its page \footnote{https://github.com/allenai/mice}, where we also got the code for our experiments. The only addition that was made to this code is integrating our data as an input to generate counterfactuals at each step.

\paragraph{Polyjuice} We use Polyjuice through this module\footnote{https://github.com/tongshuangwu/polyjuice}. For the generation procedure, we searched in all the control codes ('resemantic', 'restructure', 'negation', 'insert', 'lexical', 'shuffle', 'quantifier', 'delete'), and we produce as many perturbations as it is possible for each instance. We did this by setting the $num\_perturbations = 1000$. In none of our experiments, Polyjuice had returned this plethora of results.

\paragraph{TextFooler} We utilised TextFooler through TextAttack module\footnote{https://textattack.readthedocs.io/en/latest/}. For the sake of fair comparison, we chose the same parameters as those presented in the paper by the authors. The constraints concern the disallowing of the modification of stopwords and already modified words. Also, the threshold of the word embedding distance that two words are considered synonyms is defined as 0.5, and we force the replacements to be based on pos tagging.

\section{Newsgroups further analysis}
\label{ap:news}
\begin{figure*}[!p]
\centering
\begin{subfigure}[b]{0.33\textwidth}
\includegraphics[width=\textwidth]{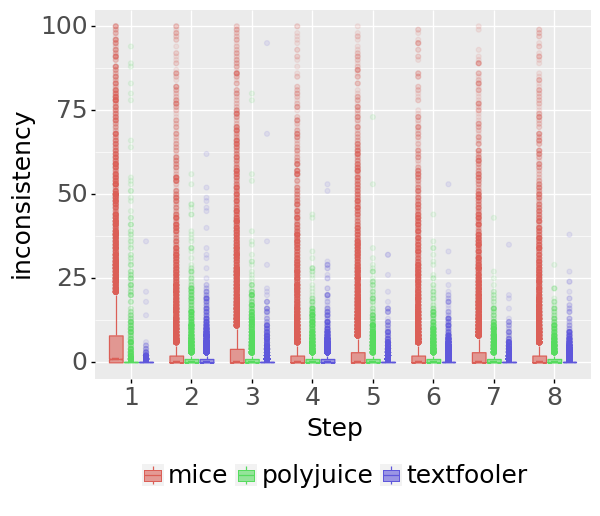}
    \caption{Inconsistency of minimality.}
    \label{fig:news-ico}
\end{subfigure}
\begin{subfigure}[b]{0.32\textwidth}
\includegraphics[width=\textwidth]{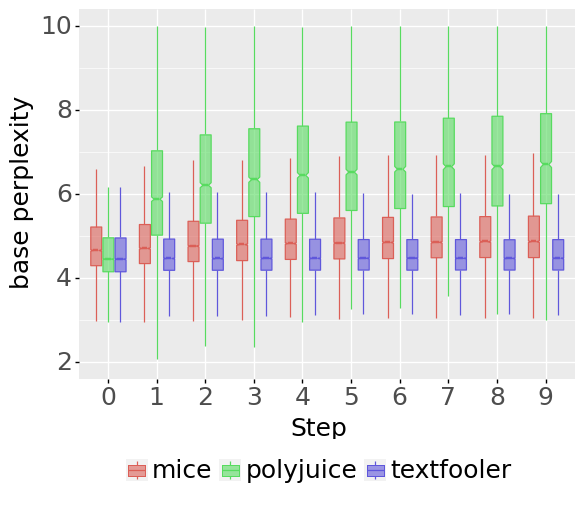}
\caption{Perplexity of base GPT-2.}
\label{fig:news-ppl}
\end{subfigure}
\begin{subfigure}[b]{0.32\textwidth}
\centering
\includegraphics[width=\textwidth]{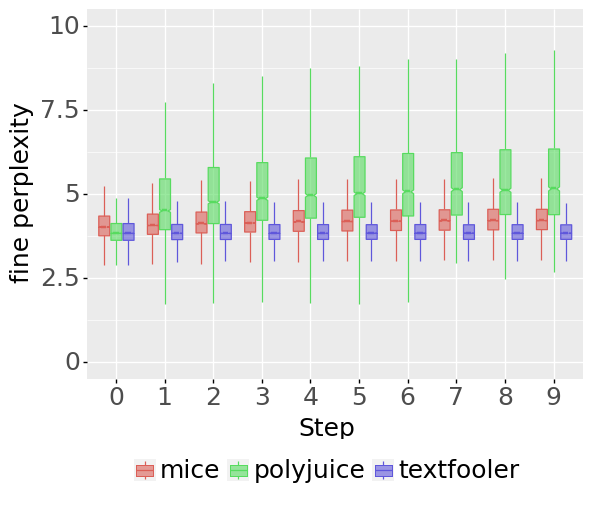}
\caption{Perplexity of fine GPT-2.}
\label{fig:news-fine}

\end{subfigure}
\caption{Inc@n, Perplexity of base GPT-2 and Perplexity of fine GPT-2 for the Newsgroups Dataset.}
\label{fig: news_figs}
\end{figure*}

Figure \ref{fig: news_figs} depicts the inconsistency of minimality, the perplexity of base GPT-2, and the perplexity of fine GPT-2 for each editor on the Newsgroups dataset. As this task is not binary, there is no pattern between odd and even steps that we observed on the IMDb dataset, but there is consistent behaviour. More specifically, since the editor does not have to return to the original class, but to any other class at each feedback step, the difficulty of flipping labels is similar between even and odd steps. In fact, if we isolate the cases where the editor returns to the original class, the behaviours observed on IMDb still hold. Furthermore, the fact that Newsgroups is a multi-class dataset, seems to make MiCE struggle more than the other editors (see also Table \ref{tab:inc}) due to the fact that MiCE requires a target class to be specified, and edits the text accordingly to flip to that class, while Polyjuice and TextFooler allow the option to perform edits just to change the class of the input, to \emph{any} other class. We address this requirement by defining as a target class for each step, the second class of the prediction, which is also the default methodology that the editor's creators follow in their study. So the task that MiCE performs is harder than the others (editing a text to be classified from class A to class B is at least as hard as editing the text to be classified from class A to any other class), and could lead to the observed higher inconsistency values for MiCE and the different behaviour compared to the IMDb dataset.  

Based on the figures, we can conclude that the proposed method produces consistent results for the behaviour of each editor even with fewer steps. We can thus significantly reduce the computational cost of the method, as just two or three steps are enough for drawing reliable conclusions.

\begin{figure}[h!]
\centering
\begin{subfigure}[b]{0.45\textwidth}
\includegraphics[width=\textwidth]{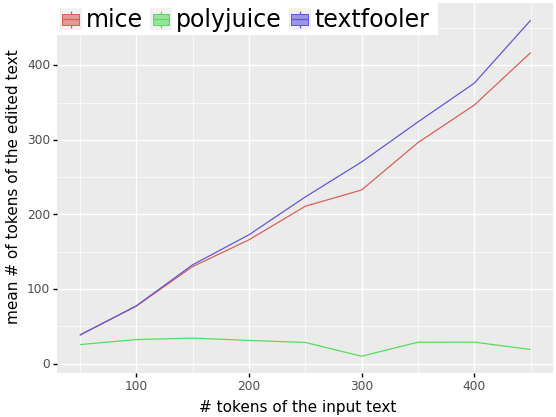}
    \caption{Mean number of tokens of the edited text regarding the number of tokens of the input of the IMDb Dataset.}
    \label{fig:imdb-prun}
\end{subfigure}

\begin{subfigure}[b]{0.45\textwidth}
\includegraphics[width=\textwidth]{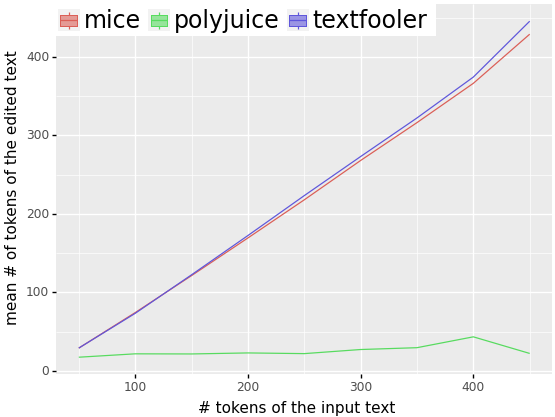}
    \caption{Mean number of tokens of the edited text regarding the number of tokens of the input of the Newsgroups Dataset.}
    \label{fig:news-prun}
\end{subfigure}

\caption{Mean number of tokens of the edited text regarding the number of tokens of the input.}
\label{fig: mean_tokens}
\end{figure}

\begin{figure}[!h]
\centering
\begin{subfigure}[b]{0.49\textwidth}
\includegraphics[width=\textwidth]{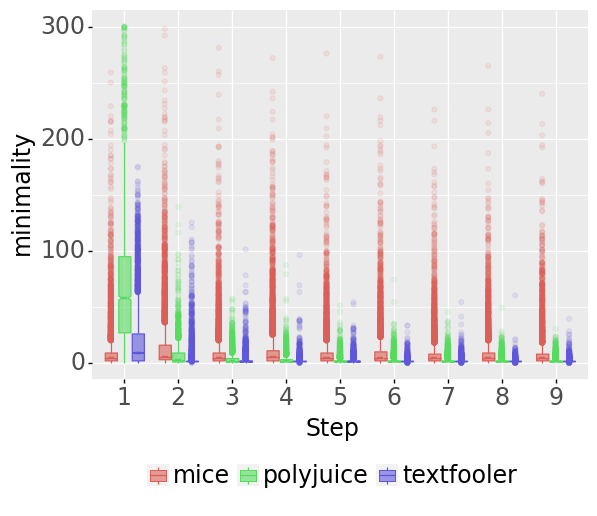}
    \caption{Minimality@n for the Newsgroups dataset.}
    \label{fig:news-minimality}
\end{subfigure}

\caption{Minimality@n for the Newsgroups Dataset.}
\label{fig: news_figs_minimality}
\end{figure}

\section{Length of Counterfactual texts}
\label{sec:prunning}

In order to further investigate the behaviour of each editor regarding the number of tokens of the input text, we present Figure \ref{fig: mean_tokens}, which depicts the mean number of tokens of the edited texts relative to the number of input tokens. 
The output texts of MiCE and Textfooler are distributed equally with the input text for both of the studied datasets. However, Polyjuice produces text with a limited length. There are also cases where the produced text of Polyjuice is longer, but they are not representative. It is worth mentioning that this may be caused due by the inner mechanism (e.g. GPT-2) of this method but also due to the evaluation method. As Polyjuice is a method for counterfactual generation, for the evaluation procedure, we preferred a text that has a different label than the original text (achieve the counterfactual goal) instead of one closest one that is classified on the same class \cite{madsen2022post}. This, combined with the task-agnostic nature of Polyjuice,  forced it to make more aggressive edits by pruning a significant portion of the original text, which is a constant behaviour along the datasets.

\section{Size of test set}
\label{sec:t_test}

In order to investigate the effect of the sample size on the results of the proposed metric, we conducted multiple t-tests for different sample sizes, feedback steps, and datasets. In particular, we selected four subsets of 10, 50, 100, and 200 samples, and we performed t-tests between the values of their inconsistencies of every pair of the editors in order to find out at which point their values are significantly different from each other. The p-values of these experiments are shown in Table \ref{tab:table_t_imdb} for the IMDb dataset and Table \ref{tab:table_t_news} for the Newsgroups. In these tables, the values that are consistently (for each feedback step) less than 0.05 are shown in bold. For every feedback step in the IMDb dataset, p is less than 0.05 for sample sizes greater than \textbf{100}, while for Newsgroups,  the same holds true for sample sizes greater than \textbf{200}. 

\label{appendix:t_test}
\begin{center}
\begin{table*}
\scalebox{0.95}{
\begin{tabular}{|ccccccccc|}
\hline
\multicolumn{9}{|c|}{MiCE}                         \\ \hline
\multicolumn{9}{|c|}{Sample Size: 10}               \\ \hline
\multicolumn{1}{|c|}{}           & \multicolumn{1}{c|}{1}               & \multicolumn{1}{c|}{2}                 & \multicolumn{1}{c|}{3}               & \multicolumn{1}{c|}{4}                 & \multicolumn{1}{c|}{5}                 & \multicolumn{1}{c|}{6}                 & \multicolumn{1}{c|}{7}                & 8                 \\ \hline
\multicolumn{1}{|c|}{Polyjuice}  & \multicolumn{1}{c|}{0.2812}          & \multicolumn{1}{c|}{0.588}             & \multicolumn{1}{c|}{0.3563}          & \multicolumn{1}{c|}{0.3219}            & \multicolumn{1}{c|}{0.1093}            & \multicolumn{1}{c|}{0.3376}            & \multicolumn{1}{c|}{0.3039}           & 0.133             \\ \hline
\multicolumn{1}{|c|}{TextFooler} & \multicolumn{1}{c|}{0.4788}          & \multicolumn{1}{c|}{0.4853}            & \multicolumn{1}{c|}{0.2538}          & \multicolumn{1}{c|}{0.2107}            & \multicolumn{1}{c|}{0.2014}            & \multicolumn{1}{c|}{0.6249}            & \multicolumn{1}{c|}{0.0695}           & 0.1658            \\ \hline
\multicolumn{9}{|c|}{Sample Size: 50}              \\ \hline
\multicolumn{1}{|c|}{Polyjuice}  & \multicolumn{1}{c|}{0.0383}          & \multicolumn{1}{c|}{0.342}             & \multicolumn{1}{c|}{0.0266}          & \multicolumn{1}{c|}{0.0073}            & \multicolumn{1}{c|}{0.1714}            & \multicolumn{1}{c|}{0.0377}            & \multicolumn{1}{c|}{0.0852}           & 0.1184            \\ \hline
\multicolumn{1}{|c|}{TextFooler} & \multicolumn{1}{c|}{0.2805}          & \multicolumn{1}{c|}{1.232e-05}         & \multicolumn{1}{c|}{0.2646}          & \multicolumn{1}{c|}{0.004}             & \multicolumn{1}{c|}{0.03}              & \multicolumn{1}{c|}{0.0054}            & \multicolumn{1}{c|}{0.1028}           & 0.0063            \\ \hline
\multicolumn{9}{|c|}{Sample Size: 100}             \\ \hline
\multicolumn{1}{|c|}{Polyjuice}  & \multicolumn{1}{c|}{\textbf{0.0252}} & \multicolumn{1}{c|}{\textbf{0.0168}}   & \multicolumn{1}{c|}{\textbf{0.0001}} & \multicolumn{1}{c|}{\textbf{0.0001}}   & \multicolumn{1}{c|}{\textbf{0.0048}}   & \multicolumn{1}{c|}{\textbf{0.0091}}   & \multicolumn{1}{c|}{\textbf{0.0081}}  & \textbf{0.0003}   \\ \hline
\multicolumn{1}{|c|}{TextFooler} & \multicolumn{1}{c|}{\textbf{0.0495}} & \multicolumn{1}{c|}{\textbf{6e-08}}    & \multicolumn{1}{c|}{\textbf{0.0104}} & \multicolumn{1}{c|}{\textbf{0.0001}}   & \multicolumn{1}{c|}{\textbf{0.0016}}   & \multicolumn{1}{c|}{\textbf{0.0003}}   & \multicolumn{1}{c|}{\textbf{0.0032}}  & \textbf{0.0001}   \\ \hline
\multicolumn{9}{|c|}{Sample Size: 200}             \\ \hline
\multicolumn{1}{|c|}{Polyjuice}  & \multicolumn{1}{c|}{\textbf{0.0084}} & \multicolumn{1}{c|}{\textbf{0.0036}}   & \multicolumn{1}{c|}{\textbf{1e-08}}  & \multicolumn{1}{c|}{\textbf{4.02e-05}} & \multicolumn{1}{c|}{\textbf{2.72e-05}} & \multicolumn{1}{c|}{\textbf{0.0012}}   & \multicolumn{1}{c|}{\textbf{0.0007}}  & \textbf{5.66e-06} \\ \hline
\multicolumn{1}{|c|}{TextFooler} & \multicolumn{1}{c|}{\textbf{0.0461}} & \multicolumn{1}{c|}{\textbf{2.73e-14}} & \multicolumn{1}{c|}{\textbf{0.0013}} & \multicolumn{1}{c|}{\textbf{1.3e-10}}  & \multicolumn{1}{c|}{\textbf{0.0006}}   & \multicolumn{1}{c|}{\textbf{4.89e-07}} & \multicolumn{1}{c|}{\textbf{2.2e-05}} & \textbf{4.2e-08}  \\ \hline

\multicolumn{9}{|c|}{Sample Size: 500}             \\ \hline
\multicolumn{1}{|c|}{Polyjuice}  & \multicolumn{1}{c|}{\textbf{0.00043}} & \multicolumn{1}{c|}{\textbf{0.0}}   & \multicolumn{1}{c|}{\textbf{0.036}}  & \multicolumn{1}{c|}{\textbf{0.0}} & \multicolumn{1}{c|}{\textbf{0.0006}} & \multicolumn{1}{c|}{\textbf{1e-08}}   & \multicolumn{1}{c|}{\textbf{0.001}}  & \textbf{0.0} \\ \hline
\multicolumn{1}{|c|}{TextFooler} & \multicolumn{1}{c|}{\textbf{0.0368}} & \multicolumn{1}{c|}{\textbf{0.0}} & \multicolumn{1}{c|}{\textbf{1.76e-06}} & \multicolumn{1}{c|}{\textbf{0.0}}  & \multicolumn{1}{c|}{\textbf{1.86e-06}}   & \multicolumn{1}{c|}{\textbf{0.0}} & \multicolumn{1}{c|}{\textbf{4.2e-05}} & \textbf{0.0}  
\\ \hline
\end{tabular}
}
\end{table*}
\end{center}
\begin{center}
\begin{table*}
\scalebox{0.95}{
\begin{tabular}{|ccccccccc|}
\hline
\multicolumn{9}{|c|}{Polyjuice}                     \\ \hline
\multicolumn{9}{|c|}{Sample Size: 10}                \\ \hline
\multicolumn{1}{|c|}{}          & \multicolumn{1}{c|}{1}               & \multicolumn{1}{c|}{2}                  & \multicolumn{1}{c|}{3}                  & \multicolumn{1}{c|}{4}                  & \multicolumn{1}{c|}{5}                  & \multicolumn{1}{c|}{6}               & \multicolumn{1}{c|}{7}                & \multicolumn{1}{c|}{8}                  \\ \hline
\multicolumn{1}{|c|}{mice}       & \multicolumn{1}{c|}{0.2812}          & \multicolumn{1}{c|}{0.588}              & \multicolumn{1}{c|}{0.3563}             & \multicolumn{1}{c|}{0.3219}             & \multicolumn{1}{c|}{0.1093}             & \multicolumn{1}{c|}{0.3376}          & \multicolumn{1}{c|}{0.3039}           & \multicolumn{1}{c|}{0.133}              \\ \hline
\multicolumn{1}{|c|}{textfooler} & \multicolumn{1}{c|}{0.2831}          & \multicolumn{1}{c|}{0.3649}             & \multicolumn{1}{c|}{0.3056}             & \multicolumn{1}{c|}{0.2198}             & \multicolumn{1}{c|}{0.073}              & \multicolumn{1}{c|}{0.3337}          & \multicolumn{1}{c|}{0.1573}           & \multicolumn{1}{c|}{0.047}              \\ \hline
\multicolumn{9}{|c|}{Sample Size: 50}               \\ \hline
\multicolumn{1}{|c|}{mice}       & \multicolumn{1}{c|}{0.0383}          & \multicolumn{1}{c|}{0.342}              & \multicolumn{1}{c|}{0.0266}             & \multicolumn{1}{c|}{0.0073}             & \multicolumn{1}{c|}{0.1714}             & \multicolumn{1}{c|}{0.0377}          & \multicolumn{1}{c|}{0.0852}           & \multicolumn{1}{c|}{0.1184}             \\ \hline
\multicolumn{1}{|c|}{textfooler} & \multicolumn{1}{c|}{0.0378}          & \multicolumn{1}{c|}{0.0015}             & \multicolumn{1}{c|}{0.021}              & \multicolumn{1}{c|}{0.0011}             & \multicolumn{1}{c|}{0.11}               & \multicolumn{1}{c|}{0.021}           & \multicolumn{1}{c|}{0.0199}           & \multicolumn{1}{c|}{0.0261}             \\ \hline
\multicolumn{9}{|c|}{Sample Size: 100}              \\ \hline
\multicolumn{1}{|c|}{mice}       & \multicolumn{1}{c|}{\textbf{0.0252}} & \multicolumn{1}{c|}{\textbf{0.0168}}    & \multicolumn{1}{c|}{\textbf{0.0001}}    & \multicolumn{1}{c|}{\textbf{0.0001}}    & \multicolumn{1}{c|}{\textbf{0.0048}}    & \multicolumn{1}{c|}{\textbf{0.0091}} & \multicolumn{1}{c|}{\textbf{0.0081}}  & \multicolumn{1}{c|}{\textbf{0.0003}}    \\ \hline
\multicolumn{1}{|c|}{textfooler} & \multicolumn{1}{c|}{\textbf{0.0246}} & \multicolumn{1}{c|}{\textbf{4.733e-05}} & \multicolumn{1}{c|}{\textbf{4.351e-05}} & \multicolumn{1}{c|}{\textbf{9e-06}}     & \multicolumn{1}{c|}{\textbf{0.0026}}    & \multicolumn{1}{c|}{\textbf{0.0038}} & \multicolumn{1}{c|}{\textbf{0.0003}}  & \multicolumn{1}{c|}{\textbf{4.258e-05}} \\ \hline
\multicolumn{9}{|c|}{Sample Size: 200}             \\ \hline
\multicolumn{1}{|c|}{mice}       & \multicolumn{1}{c|}{\textbf{0.0084}} & \multicolumn{1}{c|}{\textbf{0.0036}}    & \multicolumn{1}{c|}{\textbf{1e-08}}     & \multicolumn{1}{c|}{\textbf{4.028e-05}} & \multicolumn{1}{c|}{\textbf{2.723e-05}} & \multicolumn{1}{c|}{\textbf{0.0012}} & \multicolumn{1}{c|}{\textbf{0.0007}}  & \multicolumn{1}{c|}{\textbf{5.66e-06}}  \\ \hline
\multicolumn{1}{|c|}{textfooler} & \multicolumn{1}{c|}{\textbf{0.0082}} & \multicolumn{1}{c|}{\textbf{0.0}}       & \multicolumn{1}{c|}{\textbf{0.0}}       & \multicolumn{1}{c|}{\textbf{6e-08}}     & \multicolumn{1}{c|}{\textbf{1.367e-05}} & \multicolumn{1}{c|}{\textbf{0.0002}} & \multicolumn{1}{c|}{\textbf{1.6e-07}} & \multicolumn{1}{c|}{\textbf{1.3e-07}}   \\ \hline
\multicolumn{9}{|c|}{Sample Size: 500}             \\ \hline
\multicolumn{1}{|c|}{mice}       & \multicolumn{1}{c|}{\textbf{0.0004}} & \multicolumn{1}{c|}{\textbf{0.0}}    & \multicolumn{1}{c|}{\textbf{0.036}}     & \multicolumn{1}{c|}{\textbf{0.0}} & \multicolumn{1}{c|}{\textbf{0.0007}} & \multicolumn{1}{c|}{\textbf{1e-08}} & \multicolumn{1}{c|}{\textbf{0.0011}}  & \multicolumn{1}{c|}{\textbf{0.0}}
\\ \hline
\multicolumn{1}{|c|}{textfooler} & \multicolumn{1}{c|}{\textbf{2.2e-05}} & \multicolumn{1}{c|}{\textbf{0.0001}}       & \multicolumn{1}{c|}{\textbf{3.88e-05}}       & \multicolumn{1}{c|}{\textbf{0.0016}}     & \multicolumn{1}{c|}{\textbf{0.0058}} & \multicolumn{1}{c|}{\textbf{0.0009}} & \multicolumn{1}{c|}{\textbf{0.0192}} & \multicolumn{1}{c|}{\textbf{0.0007}}   \\ \hline
\end{tabular}
}
\end{table*}
\end{center}
\begin{center}
\begin{table*}
\scalebox{0.95}{
\begin{tabular}{|ccccccccc|}
\hline
\multicolumn{9}{|c|}{TextFooler}                   \\ \hline
\multicolumn{9}{|c|}{Sample Size: 10}               \\ \hline
\multicolumn{1}{|c|}{}          & \multicolumn{1}{c|}{1}               & \multicolumn{1}{c|}{2}                  & \multicolumn{1}{c|}{3}                  & \multicolumn{1}{c|}{4}               & \multicolumn{1}{c|}{5}                  & \multicolumn{1}{c|}{6}                & \multicolumn{1}{c|}{7}                  & 8                  \\ \hline
\multicolumn{1}{|c|}{polyjuice} & \multicolumn{1}{c|}{0.2831}          & \multicolumn{1}{c|}{0.3649}             & \multicolumn{1}{c|}{0.3056}             & \multicolumn{1}{c|}{0.2198}          & \multicolumn{1}{c|}{0.073}              & \multicolumn{1}{c|}{0.3337}           & \multicolumn{1}{c|}{0.1573}             & 0.047              \\ \hline
\multicolumn{1}{|c|}{mice}      & \multicolumn{1}{c|}{0.4788}          & \multicolumn{1}{c|}{0.4853}             & \multicolumn{1}{c|}{0.2538}             & \multicolumn{1}{c|}{0.2107}          & \multicolumn{1}{c|}{0.2014}             & \multicolumn{1}{c|}{0.6249}           & \multicolumn{1}{c|}{0.0695}             & 0.1658             \\ \hline
\multicolumn{9}{|c|}{Sample Size: 50}              \\ \hline
\multicolumn{1}{|c|}{polyjuice} & \multicolumn{1}{c|}{0.0378}          & \multicolumn{1}{c|}{0.0015}             & \multicolumn{1}{c|}{0.021}              & \multicolumn{1}{c|}{0.0011}          & \multicolumn{1}{c|}{0.11}               & \multicolumn{1}{c|}{0.021}            & \multicolumn{1}{c|}{0.0199}             & 0.0261             \\ \hline
\multicolumn{1}{|c|}{mice}      & \multicolumn{1}{c|}{0.2805}          & \multicolumn{1}{c|}{1.232e-05}          & \multicolumn{1}{c|}{0.2646}             & \multicolumn{1}{c|}{0.004}           & \multicolumn{1}{c|}{0.03}               & \multicolumn{1}{c|}{0.0054}           & \multicolumn{1}{c|}{0.1028}             & 0.0063             \\ \hline
\multicolumn{9}{|c|}{Sample Size: 100}             \\ \hline
\multicolumn{1}{|c|}{polyjuice} & \multicolumn{1}{c|}{\textbf{0.0246}} & \multicolumn{1}{c|}{\textbf{4.733e-05}} & \multicolumn{1}{c|}{\textbf{4.351e-05}} & \multicolumn{1}{c|}{\textbf{9e-06}}  & \multicolumn{1}{c|}{\textbf{0.0026}}    & \multicolumn{1}{c|}{\textbf{0.0038}}  & \multicolumn{1}{c|}{\textbf{0.0003}}    & \textbf{4.258e-05} \\ \hline
\multicolumn{1}{|c|}{mice}      & \multicolumn{1}{c|}{\textbf{0.0495}} & \multicolumn{1}{c|}{\textbf{6e-08}}     & \multicolumn{1}{c|}{\textbf{0.0104}}    & \multicolumn{1}{c|}{\textbf{0.0001}} & \multicolumn{1}{c|}{\textbf{0.0016}}    & \multicolumn{1}{c|}{\textbf{0.0003}}  & \multicolumn{1}{c|}{\textbf{0.0032}}    & \textbf{0.0001}    \\ \hline
\multicolumn{9}{|c|}{Sample Size: 200}              \\ \hline
\multicolumn{1}{|c|}{polyjuice} & \multicolumn{1}{c|}{\textbf{0.0082}} & \multicolumn{1}{c|}{\textbf{0.0}}       & \multicolumn{1}{c|}{\textbf{0.0}}       & \multicolumn{1}{c|}{\textbf{6e-08}}  & \multicolumn{1}{c|}{\textbf{1.367e-05}} & \multicolumn{1}{c|}{\textbf{0.0002}}  & \multicolumn{1}{c|}{\textbf{1.6e-07}}   & \textbf{1.3e-07}   \\ \hline
\multicolumn{1}{|c|}{mice}      & \multicolumn{1}{c|}{\textbf{0.0461}} & \multicolumn{1}{c|}{\textbf{0.0}}       & \multicolumn{1}{c|}{\textbf{0.0013}}    & \multicolumn{1}{c|}{\textbf{0.0}}    & \multicolumn{1}{c|}{\textbf{0.0006}}    & \multicolumn{1}{c|}{\textbf{4.9e-07}} & \multicolumn{1}{c|}{\textbf{2.225e-05}} & \textbf{4e-08}     \\ \hline
\multicolumn{9}{|c|}{Sample Size: 500}              \\ \hline
\multicolumn{1}{|c|}{polyjuice} & \multicolumn{1}{c|}{\textbf{2.28e-05}} & \multicolumn{1}{c|}{\textbf{0.0001}}       & \multicolumn{1}{c|}{\textbf{3.882e-05}}       & \multicolumn{1}{c|}{\textbf{0.0016}}  & \multicolumn{1}{c|}{\textbf{0.0058}} & \multicolumn{1}{c|}{\textbf{0.0009}}  & \multicolumn{1}{c|}{\textbf{0.0192}}   & \textbf{0.0007}   \\ \hline
\multicolumn{1}{|c|}{mice}      & \multicolumn{1}{c|}{\textbf{0.0369}} & \multicolumn{1}{c|}{\textbf{0.0}}       & \multicolumn{1}{c|}{\textbf{1.76e-06}}    & \multicolumn{1}{c|}{\textbf{0.0}}    & \multicolumn{1}{c|}{\textbf{1.86e-06}}    & \multicolumn{1}{c|}{\textbf{0.0}} & \multicolumn{1}{c|}{\textbf{4.251e-05}} & \textbf{0.0}     \\ \hline

\end{tabular}
}
\caption{P-value of the inconsistency of different sample sizes of the IMDb dataset.}
\label{tab:table_t_imdb}
\end{table*}
\end{center}

\begin{table*}[]
\scalebox{0.95}{
\begin{tabular}{|ccccccccc|}
\hline
\multicolumn{9}{|c|}{MiCE}                         \\ \hline
\multicolumn{9}{|c|}{Sample Size: 10}              \\ \hline
\multicolumn{1}{|c|}{polyjuice}        & \multicolumn{1}{c|}{0.1311}             & \multicolumn{1}{c|}{0.0963}          & \multicolumn{1}{c|}{0.4378}             & \multicolumn{1}{c|}{0.0042}          & \multicolumn{1}{c|}{0.1069}            & \multicolumn{1}{c|}{0.2044}          & \multicolumn{1}{c|}{0.1384}          & 0.1809             \\ \hline
\multicolumn{1}{|c|}{textfooler}             & \multicolumn{1}{c|}{0.0717}             & \multicolumn{1}{c|}{0.2283}          & \multicolumn{1}{c|}{0.0924}             & \multicolumn{1}{c|}{0.3264}          & \multicolumn{1}{c|}{0.165}             & \multicolumn{1}{c|}{0.2194}          & \multicolumn{1}{c|}{0.5411}          & 0.0453             \\ \hline
\multicolumn{9}{|c|}{Sample Size: 50}              \\ \hline
\multicolumn{1}{|c|}{polyjuice}    &
\multicolumn{1}{c|}{0.1311}             & \multicolumn{1}{c|}{0.0963}          & \multicolumn{1}{c|}{0.4378}             & \multicolumn{1}{c|}{0.0042}          & \multicolumn{1}{c|}{0.1069}            & \multicolumn{1}{c|}{0.2044}          & \multicolumn{1}{c|}{0.1384}          & 0.1809 
\\ \hline
\multicolumn{1}{|c|}{textfooler}             & \multicolumn{1}{c|}{0.0006}             & \multicolumn{1}{c|}{0.0064}          & \multicolumn{1}{c|}{0.0338}             & \multicolumn{1}{c|}{0.1265}          & \multicolumn{1}{c|}{0.008}             & \multicolumn{1}{c|}{0.1015}          & \multicolumn{1}{c|}{0.0995}          & 0.0101             \\ \hline
\multicolumn{9}{|c|}{Sample Size: 100}             \\ \hline
\multicolumn{1}{|c|}{polyjuice}  &
\multicolumn{1}{c|}{0.0033}             & \multicolumn{1}{c|}{2.64e-06}        & \multicolumn{1}{c|}{0.0177}             & \multicolumn{1}{c|}{1e-08}           & \multicolumn{1}{c|}{0.0017}            & \multicolumn{1}{c|}{2.878e-05}       & \multicolumn{1}{c|}{0.081}           & 0.0049 
\\ \hline
\multicolumn{1}{|c|}{textfooler}             & \multicolumn{1}{c|}{\textbf{1.29e-06}}  & \multicolumn{1}{c|}{\textbf{0.0004}} & \multicolumn{1}{c|}{\textbf{0.0034}}    & \multicolumn{1}{c|}{\textbf{0.0001}} & \multicolumn{1}{c|}{\textbf{0.0009}}   & \multicolumn{1}{c|}{\textbf{0.0104}} & \multicolumn{1}{c|}{\textbf{0.0344}} & \textbf{0.0007}    \\ \hline
\multicolumn{9}{|c|}{Sample Size: 200} \\ \hline
\multicolumn{1}{|c|}{polyjuice}             & \multicolumn{1}{c|}{\textbf{0.0}}       & \multicolumn{1}{c|}{\textbf{0.0005}} & \multicolumn{1}{c|}{\textbf{4.937e-05}} & \multicolumn{1}{c|}{\textbf{0.0002}} & \multicolumn{1}{c|}{\textbf{2.13e-06}} & \multicolumn{1}{c|}{\textbf{0.0003}} & \multicolumn{1}{c|}{\textbf{0.006}}  & \textbf{0.0041}    \\ \hline
\multicolumn{1}{|c|}{textfooler}       & \multicolumn{1}{c|}{\textbf{1.513e-05}} & \multicolumn{1}{c|}{\textbf{0.0}}    & \multicolumn{1}{c|}{\textbf{2.8e-07}}   & \multicolumn{1}{c|}{\textbf{0.0}}    & \multicolumn{1}{c|}{\textbf{3.5e-07}}  & \multicolumn{1}{c|}{\textbf{0.0}}    & \multicolumn{1}{c|}{\textbf{0.0043}} & \textbf{2e-08}     \\ \hline
\multicolumn{9}{|c|}{Sample Size: 500} \\ \hline
\multicolumn{1}{|c|}{polyjuice}             & \multicolumn{1}{c|}{\textbf{0.0}} &
\multicolumn{1}{c|}{\textbf{2.32e-06}} &
\multicolumn{1}{c|}{\textbf{0.0}} &
\multicolumn{1}{c|}{\textbf{1.09e-06}} &
\multicolumn{1}{c|}{\textbf{0.0}} &
\multicolumn{1}{c|}{\textbf{5e-08}} &
\multicolumn{1}{c|}{\textbf{0.0}} &
\multicolumn{1}{c|}{\textbf{1e-08}}
\\ \hline
\multicolumn{1}{|c|}{textfooler}   & \multicolumn{1}{c|}{\textbf{0.0}} &
\multicolumn{1}{c|}{\textbf{3.3e-07}} &
\multicolumn{1}{c|}{\textbf{0.0}} &
\multicolumn{1}{c|}{\textbf{3e-08}} &
\multicolumn{1}{c|}{\textbf{0.0}} &
\multicolumn{1}{c|}{\textbf{0.0}} &
\multicolumn{1}{c|}{\textbf{0.0}} &
\multicolumn{1}{c|}{\textbf{0.0}}     \\ \hline

\end{tabular}
}
\end{table*}

\begin{table*}[]
\scalebox{0.95}{
\begin{tabular}{|ccccccccc|}
\hline
\multicolumn{9}{|c|}{Polyjuice}                   \\ \hline
\multicolumn{9}{|c|}{Sample Size: 10}              \\ \hline
\multicolumn{1}{|c|}{}           & \multicolumn{1}{c|}{1}                  & \multicolumn{1}{c|}{2}            & \multicolumn{1}{c|}{3}                & \multicolumn{1}{c|}{4}            & \multicolumn{1}{c|}{5}                & \multicolumn{1}{c|}{6}               & \multicolumn{1}{c|}{7}               & 8               \\ \hline
\multicolumn{1}{|c|}{mice}       & \multicolumn{1}{c|}{0.3306}             & \multicolumn{1}{c|}{0.0995}       & \multicolumn{1}{c|}{0.1407}           & \multicolumn{1}{c|}{0.0043}       & \multicolumn{1}{c|}{0.7421}           & \multicolumn{1}{c|}{0.4028}          & \multicolumn{1}{c|}{0.1387}          & 0.2846          \\ \hline
\multicolumn{1}{|c|}{textfooler} & \multicolumn{1}{c|}{0.1311}             & \multicolumn{1}{c|}{0.0963}       & \multicolumn{1}{c|}{0.4378}           & \multicolumn{1}{c|}{0.0042}       & \multicolumn{1}{c|}{0.1069}           & \multicolumn{1}{c|}{0.2044}          & \multicolumn{1}{c|}{0.1384}          & 0.1809          \\ \hline
\multicolumn{9}{|c|}{Sample Size: 50}              \\ \hline
\multicolumn{1}{|c|}{mice}       & \multicolumn{1}{c|}{0.9654}             & \multicolumn{1}{c|}{0.0065}       & \multicolumn{1}{c|}{0.6376}           & \multicolumn{1}{c|}{8e-08}        & \multicolumn{1}{c|}{0.7342}           & \multicolumn{1}{c|}{0.3373}          & \multicolumn{1}{c|}{0.1692}          & 0.9168          \\ \hline
\multicolumn{1}{|c|}{textfooler} & \multicolumn{1}{c|}{0.0033}             & \multicolumn{1}{c|}{2.64e-06}     & \multicolumn{1}{c|}{0.0177}           & \multicolumn{1}{c|}{1e-08}        & \multicolumn{1}{c|}{0.0017}           & \multicolumn{1}{c|}{2.878e-05}       & \multicolumn{1}{c|}{0.081}           & 0.0049          \\ \hline
\multicolumn{9}{|c|}{Sample Size: 100}             \\ \hline
\multicolumn{1}{|c|}{mice}       & \multicolumn{1}{c|}{0.3659}             & \multicolumn{1}{c|}{1e-08}        & \multicolumn{1}{c|}{0.719}            & \multicolumn{1}{c|}{0.0}          & \multicolumn{1}{c|}{0.5959}           & \multicolumn{1}{c|}{0.173}           & \multicolumn{1}{c|}{0.1055}          & 0.7947          \\ \hline
\multicolumn{1}{|c|}{textfooler} & \multicolumn{1}{c|}{0.0004}             & \multicolumn{1}{c|}{0.0}          & \multicolumn{1}{c|}{0.0002}           & \multicolumn{1}{c|}{0.0}          & \multicolumn{1}{c|}{0.0001}           & \multicolumn{1}{c|}{1e-08}           & \multicolumn{1}{c|}{0.0358}          & 4.421e-05       \\ \hline
\multicolumn{9}{|c|}{Sample Size: 200}            \\ \hline
\multicolumn{1}{|c|}{mice}       & \multicolumn{1}{c|}{\textbf{0.0468}}    & \multicolumn{1}{c|}{\textbf{0.0}} & \multicolumn{1}{c|}{\textbf{0.5025}}  & \multicolumn{1}{c|}{\textbf{0.0}} & \multicolumn{1}{c|}{\textbf{0.8023}}  & \multicolumn{1}{c|}{\textbf{0.0176}} & \multicolumn{1}{c|}{\textbf{0.0384}} & \textbf{0.2746} \\ \hline
\multicolumn{1}{|c|}{textfooler} & \multicolumn{1}{c|}{\textbf{1.513e-05}} & \multicolumn{1}{c|}{\textbf{0.0}} & \multicolumn{1}{c|}{\textbf{2.8e-07}} & \multicolumn{1}{c|}{\textbf{0.0}} & \multicolumn{1}{c|}{\textbf{3.5e-07}} & \multicolumn{1}{c|}{\textbf{0.0}}    & \multicolumn{1}{c|}{\textbf{0.0043}} & \textbf{2e-08}  \\ \hline
\multicolumn{9}{|c|}{Sample Size: 500}            \\ \hline
\multicolumn{1}{|c|}{mice}       & \multicolumn{1}{c|}{\textbf{0.0}} &
\multicolumn{1}{c|}{\textbf{2.32e-06}} &
\multicolumn{1}{c|}{\textbf{0.0}} &
\multicolumn{1}{c|}{\textbf{1.09e-06}} &
\multicolumn{1}{c|}{\textbf{0.0}} &
\multicolumn{1}{c|}{\textbf{5e-08}} &
\multicolumn{1}{c|}{\textbf{0.0}} &
\multicolumn{1}{c|}{\textbf{1e-08}}
\\ \hline
\multicolumn{1}{|c|}{textfooler} & \multicolumn{1}{c|}{\textbf{0.0005}} &
\multicolumn{1}{c|}{\textbf{0.026}} &
\multicolumn{1}{c|}{\textbf{5.3e-07}} &
\multicolumn{1}{c|}{\textbf{0.0643}} &
\multicolumn{1}{c|}{\textbf{1.8e-07}} &
\multicolumn{1}{c|}{\textbf{0.0037}} &
\multicolumn{1}{c|}{\textbf{0.0}} &
\multicolumn{1}{c|}{\textbf{0.0019}}  \\ \hline
\end{tabular}
}
\end{table*}

\begin{table*}[]
\scalebox{0.95}{
\begin{tabular}{|ccccccccc|}
\hline
\multicolumn{9}{|c|}{TextFooler}                   \\ \hline
\multicolumn{9}{|c|}{Sample Size: 10}              \\ \hline
\multicolumn{1}{|c|}{}          & \multicolumn{1}{c|}{1}                  & \multicolumn{1}{c|}{2}               & \multicolumn{1}{c|}{3}                  & \multicolumn{1}{c|}{4}               & \multicolumn{1}{c|}{5}                 & \multicolumn{1}{c|}{6}               & \multicolumn{1}{c|}{7}               & 8                  \\ \hline
\multicolumn{1}{|c|}{polyjuice} & \multicolumn{1}{c|}{0.1311}             & \multicolumn{1}{c|}{0.0963}          & \multicolumn{1}{c|}{0.4378}             & \multicolumn{1}{c|}{0.0042}          & \multicolumn{1}{c|}{0.1069}            & \multicolumn{1}{c|}{0.2044}          & \multicolumn{1}{c|}{0.1384}          & 0.1809             \\ \hline
\multicolumn{1}{|c|}{mice}      & \multicolumn{1}{c|}{0.0717}             & \multicolumn{1}{c|}{0.2283}          & \multicolumn{1}{c|}{0.0924}             & \multicolumn{1}{c|}{0.3264}          & \multicolumn{1}{c|}{0.165}             & \multicolumn{1}{c|}{0.2194}          & \multicolumn{1}{c|}{0.5411}          & 0.0453             \\ \hline
\multicolumn{9}{|c|}{Sample Size: 50}              \\ \hline
\multicolumn{1}{|c|}{polyjuice} & \multicolumn{1}{c|}{0.0033}             & \multicolumn{1}{c|}{2.64e-06}        & \multicolumn{1}{c|}{0.0177}             & \multicolumn{1}{c|}{1e-08}           & \multicolumn{1}{c|}{0.0017}            & \multicolumn{1}{c|}{2.878e-05}       & \multicolumn{1}{c|}{0.081}           & 0.0049             \\ \hline
\multicolumn{1}{|c|}{mice}      & \multicolumn{1}{c|}{0.0006}             & \multicolumn{1}{c|}{0.0064}          & \multicolumn{1}{c|}{0.0338}             & \multicolumn{1}{c|}{0.1265}          & \multicolumn{1}{c|}{0.008}             & \multicolumn{1}{c|}{0.1015}          & \multicolumn{1}{c|}{0.0995}          & 0.0101             \\ \hline
\multicolumn{9}{|c|}{Sample Size: 100}             \\ \hline
\multicolumn{1}{|c|}{polyjuice} & \multicolumn{1}{c|}{\textbf{0.0004}}    & \multicolumn{1}{c|}{\textbf{0.0}}    & \multicolumn{1}{c|}{\textbf{0.0002}}    & \multicolumn{1}{c|}{\textbf{0.0}}    & \multicolumn{1}{c|}{\textbf{0.0001}}   & \multicolumn{1}{c|}{\textbf{1e-08}}  & \multicolumn{1}{c|}{\textbf{0.0358}} & \textbf{4.421e-05} \\ \hline
\multicolumn{1}{|c|}{mice}      & \multicolumn{1}{c|}{\textbf{1.29e-06}}  & \multicolumn{1}{c|}{\textbf{0.0004}} & \multicolumn{1}{c|}{\textbf{0.0034}}    & \multicolumn{1}{c|}{\textbf{0.0001}} & \multicolumn{1}{c|}{\textbf{0.0009}}   & \multicolumn{1}{c|}{\textbf{0.0104}} & \multicolumn{1}{c|}{\textbf{0.0344}} & \textbf{0.0007}    \\ \hline
\multicolumn{9}{|c|}{Sample Size: 200}             \\ \hline
\multicolumn{1}{|c|}{polyjuice} & \multicolumn{1}{c|}{\textbf{1.513e-05}} & \multicolumn{1}{c|}{\textbf{0.0}}    & \multicolumn{1}{c|}{\textbf{2.8e-07}}   & \multicolumn{1}{c|}{\textbf{0.0}}    & \multicolumn{1}{c|}{\textbf{3.5e-07}}  & \multicolumn{1}{c|}{\textbf{0.0}}    & \multicolumn{1}{c|}{\textbf{0.0043}} & \textbf{2e-08}     \\ \hline
\multicolumn{1}{|c|}{mice}      & \multicolumn{1}{c|}{\textbf{0.0}}       & \multicolumn{1}{c|}{\textbf{0.0005}} & \multicolumn{1}{c|}{\textbf{4.937e-05}} & \multicolumn{1}{c|}{\textbf{0.0002}} & \multicolumn{1}{c|}{\textbf{2.13e-06}} & \multicolumn{1}{c|}{\textbf{0.0003}} & \multicolumn{1}{c|}{\textbf{0.006}}  & \textbf{0.0041}    \\ \hline
\multicolumn{9}{|c|}{Sample Size: 500}             \\ \hline
\multicolumn{1}{|c|}{polyjuice} & \multicolumn{1}{c|}{\textbf{0.0005}} &
\multicolumn{1}{c|}{\textbf{0.0213}} &
\multicolumn{1}{c|}{\textbf{5.3e-07}} &
\multicolumn{1}{c|}{\textbf{0.0643}} &
\multicolumn{1}{c|}{\textbf{1.8e-07}} &
\multicolumn{1}{c|}{\textbf{0.0037}} &
\multicolumn{1}{c|}{\textbf{0.0}} &
\multicolumn{1}{c|}{\textbf{0.0019}}     \\ \hline
\multicolumn{1}{|c|}{mice}      & \multicolumn{1}{c|}{\textbf{0.0}} &
\multicolumn{1}{c|}{\textbf{3.3e-07}} &
\multicolumn{1}{c|}{\textbf{0.0}} &
\multicolumn{1}{c|}{\textbf{3e-08}} &
\multicolumn{1}{c|}{\textbf{0.0}} &
\multicolumn{1}{c|}{\textbf{0.0}} &
\multicolumn{1}{c|}{\textbf{0.0}} &
\multicolumn{1}{c|}{\textbf{0.0}}    \\ \hline
\end{tabular}
}
\caption{P-value of the inconsistency of different sample sizes of the Newsgroup dataset.}
\label{tab:table_t_news}
\end{table*}



\end{document}